# Linguistic Information Energy

James Ford
jford@CACI.com

ABSTRACT

In this treatment a text is considered to be a series of word impulses which are read at a constant rate. The brain then assembles these units of information into higher units of meaning. A classical systems approach is used to model an initial part of this assembly process. The concepts of linguistic system response, information energy, and ordering energy are defined and analyzed. Finally, as a demonstration, information energy is used to estimate the publication dates of a series of texts and the similarity of a set of texts.

## Introduction

This paper represents an effort to apply a classic systems approach to the modeling of linguistic processes. Specifically a model is posited from which statistics about the effort required by the reader of a given text can be drawn. This effort is termed the information energy.

The motivational linguistic model is developed first. This is followed by several computations exploring typical behavior of the statistic.

## Linguistic Model

**Definitions.** To begin, let the impulsive value of a word, $w$, be given by $\text{I}(w)$. This value should represent the work required by the reader in recognizing a word. Words that are rare or long or are spelt in an unPracticed manner should have a higher value. Some examples of $\text{I}(w)$ are given in Table 1. Here $\Pr(w)$ is the probability of $w$ and $l(w)$ is a measure of the length of $w$ in characters or syllables or other lexical units. These functions will be referred to later.

Table 1. Word Impulse Functions

| Name | Formula |
|---|---|
| $\text{I}_1(w)$ | $\log(l(w)/\Pr(w))$ |
| $\text{I}_2(w)$ | $-l(w) \cdot \log(\Pr(w))$ |
| $\text{I}_3(w)$ | $-\log(\Pr(w))$ |
| $\text{I}_4(w)$ | $-\log(l(w)) * \log(\Pr(w))$ |
| $\text{I}_5(w)$ | $l(w)$ |
| $\text{I}_6(w)$ | $\log(l(w))$ |

$I_3(w)$ is classically the amount of information resolved by the occurrence of word $w$.

Now a text can be represented as a time series of word values
$$Text(t) = \sum_i I(w_i) \cdot \delta(i-t)$$
where $\delta(\cdot)$ is the Dirac delta. Note that the time variable $t$ is continuous while $i$ is discrete.

Next, define a memory smoothing function (MSF) or forgetting function, $M_\tau(t)$, which represents the fact that a reader doesn't need to remember all the words previously read to extract meaning, only the most recent ones. $M_\tau(t)$ is positive and non-increasing. The subscript $\tau$ is the width of the interval over which the MSF is active.

As an example of an MSF consider
$$M_{\tau,k}(t) = P_\tau(t) + U(\tau+1-t) \cdot \exp[-(\tau+1-t) \cdot k] \qquad (1)$$
where $P_\tau(t) = 1$ on the interval $[0, \tau+1)$ and is zero elsewhere and $U(\tau - t) = 1$ for $\tau > t$ and is zero elsewhere. A special case of this, $M_{2,\infty}(t)$, will be used in the following examples.

A finite value of k may best represent real linguistic processing but in practice any $\tau, k$ pair seems to have a corresponding $\tau_{eq}, \infty$ pair that gives equivalent performance.

$M_\tau(t)$ is the linguistic system response (LSR) of a single word with an impulse value of 1 although in reality a reader would not continue to read after reading the only word in a single word text. The LSR for a text is then $LSR(Text, t) = M_\tau(t) * Text(t)$ where $*$ is the standard convolution operator [1].

If the LSR is the minimum attention level required to read a text at time $t$ and the energy required to maintain that attention level is equal to it's square, then the average information energy required by a finite text is
$$IE(Text) = \tfrac{1}{T} \int_0^T LSR(Text, t)^2 \, dt$$
where $T$ is the time extent of the text. Note that a more general term for this might be decoding energy since the exact choice of word impulse function $I(w)$ hasn't been specified. However, the term information energy will be used throughout this paper.

The information energy defined here is not related to the information energy of Onicescu [2].

**Example.** Consider a random text, $R$, of $N$ words with $I(w)$ uniformly distributed on $[0, D]$ and an MSF of the form $M_{2,\infty}(t)$ given by (1).

With the exception of the first word, evaluation of the LSR will be the evaluation of the sum of two random variables $x = I(w_i) + I(w_{i+1})$ and this sum will have the density shown in Figure 1.

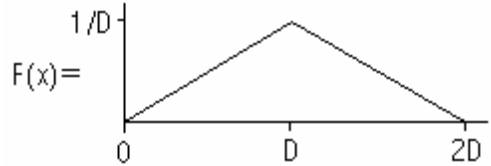

Figure 1. Probability Density for Word Impulse Sum

The expected energy for this sum of variables will be
$$E[x^2] = \int x^2 \cdot F(x) \cdot dx = \tfrac{7}{6} D^2.$$
Thus the expected information energy will be
$$E[IE(R)] = \left[ \tfrac{1}{D} \int_0^D x^2 \cdot dx + (N-1) \cdot \tfrac{7}{6} D^2 \right] / N = D^2 \left[ \tfrac{1}{3} + (N-1) \cdot \tfrac{7}{6} \right] / N$$
where the first term in the brackets represents the evaluation of the first word. Note that the termination symbol for the text is not evaluated as a separate word, the reader merely stops reading.

**Definition.** Let $O$ be an ordering applied to a text $T$ such that if $w_i, w_j \in O(T)$ and $i < j$ then $I(w_i) \geq I(w_j)$. The ordering energy of the text is defined as
$$OE(T) = IE(O(T)) - IE(T).$$

Ordering energy is identically zero for $\tau = 1$ and can be zero for $\tau > 1$ since the ordering $O$ may not change the word order. But it can also be negative. As an example consider the text $T = w_1 w_2 w_3$ with $I(w_2) > I(w_1) > I(w_3)$ then $IE(T) > IE(O(T))$ if the MSF is $M_{2,\infty}(t)$. Negative ordering energy was seen less than 10% of the time for $\tau = 2$ and decreased by half for every increase in the value of $\tau$ by 1.

In practice the ordering energy will be zero 3%-4% of the time. The largest such sentence found thus far, without repeating words, had 5 words (example: "How can they do it"). The author will NOT be offering a prize for the largest non-trivial sentence found with an ordering energy of zero. However, it would be interesting to know.

Typically ordering energy isn't used by itself. Instead, the two statistics unordered information energy *IE(T)* and ordered information energy *IE(O(T))* are handed over in raw form for further processing.

**Example.** Let $T_2$ be a text with two alternating words $w_1, w_2$ and cardinality $N$ which is even. As in the previous example, using an MSF of the form $M_{2,\infty}(t)$ given by (1), the ordering energy is

$$OE(T_2) = \frac{N-2}{N}[I(w_1) - I(w_2)]^2.$$

## Statistical Demonstration

**Datasets.** Two datasets, the American National Corpus 2nd edition (ANC) [3] and a so called plain text corpus (PTC) constructed by the author, were used to examine the information energy of English.

The PTC is a collection of texts written in a "conversational" style. There are no government reports or scientific papers. There are no spoken transcripts, although some of the texts have quoted dialog uttered by their characters. Currently there are 58 texts 8 of which are shared with the ANC where they were labeled as either fiction or non-fiction/OUP. All were written by a single author. About 75% were written after 1900. The earliest text dates from 1529. Most came from Project Gutenberg [4].

The word probabilities for the word impulse function were calculated from the non-shared parts of both the ANC and PTC.

No pre-processing occurred other than the replacement of all numbers with a number symbol. Indeed it is the alternation of what are commonly called stop words with words of high information value that give statistics such as ordering energy a grip on the text. Stemming could reduce the average length of words in a text and cause confusion between different writing styles depending upon which word impulse function was chosen.

**Computation 1.** The first computation is a calculation of the average per-sentence information energy on the first 400 sentences of each text in the PTC making a total of 23,600 sentences and 536,583 words. For this calculation, a sentence terminator was considered the last word or symbol read by the reader with a length of 1 and a probability equal to that of a sentence. The impulsive value of each word was given by the function $I_1(w)$ from Table 1 where in this case *l(w)* is the length of the word in characters.

The MSF used was $M_{\tau,\infty}(t)$ where $\tau$ ranged between 2 and 29. The relationship between the mean information energy and $\tau$ was found to be nearly linear over that range. It can be modeled by

$$IE_{mean}(\tau) = -1162 + 580.7 \cdot \tau \qquad (2)$$

Figure 2 shows a graph of $\tau$ versus the mean information energy divided by $\tau$ and illustrates the differences between (2) and the observed data. This graph peaks at a value of $\tau$ equal to 21 while the average sentence length was 22.7 words. As the value of $\tau$ nears and then passes the average sentence length the effective value of $\tau$ will become less than the actual value slowing the rate at which $IE$ can increase.

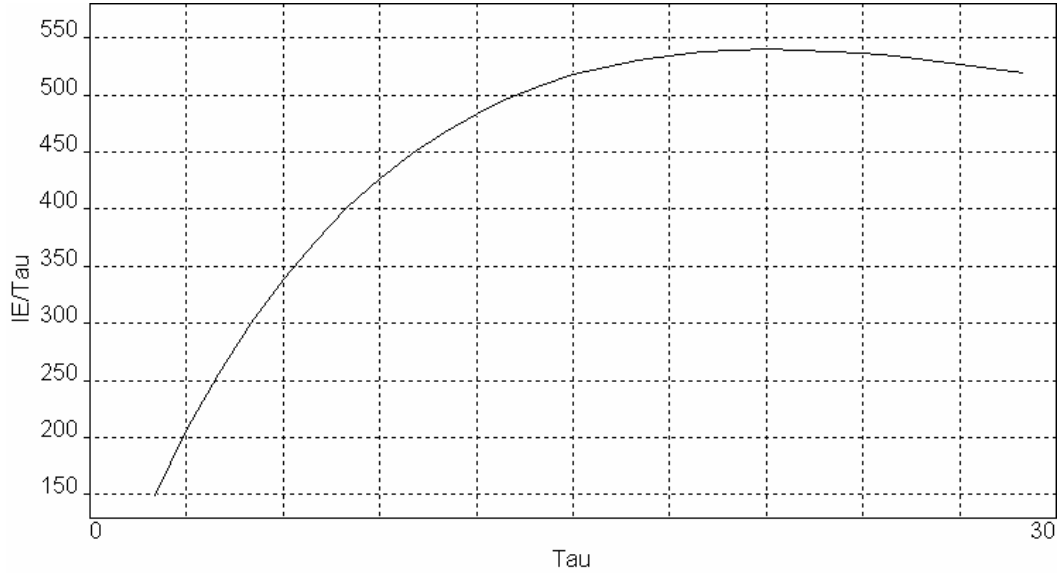

Figure 2. $\tau$ versus $IE/\tau$

Figure 3 shows the probability density estimates of the information energy when normalized (divided) by function (2).

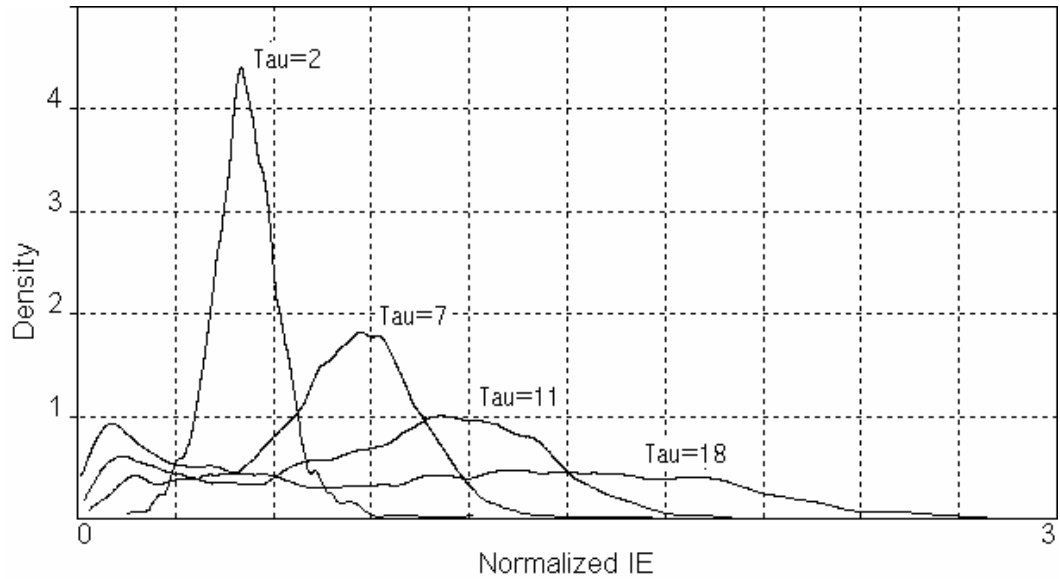

Figure 3. Normalized Information Energy Densities

The relationship found between the mean information energy, $IE_{mean}$ and the information energy's standard deviation, $IE_{SD}$, is

$$IE_{SD} = -16.4 + 0.2903 \cdot IE_{mean} + 4.4e-5 \cdot IE_{mean}^2.$$

This model is nearly exact and the 2$^{nd}$ order quality shows that something other than re-scaling of the data occurs as $\tau$ and thus $IE$ increases.

Ordering energy exhibited much the same behavior with respect to $\tau$. However, because of the large number of sentences with a zero ordering energy, 3.15% in this case, the density function will be a mixture with a spike at zero.

**Protocols.** The following two computations are focused on finding the best word impulse function from Table 1 and the best value of $\tau$ in two typical situations, estimating an abstract parameter from a text and determining the similarity of texts.

There are two way of calculating information energy that are of interest, per-sentence information energy, described in Computation 1, and per-document information energy. For the later, unordered information energy is calculated by dumping or zeroing the word impulse history in the MSF at the end of each sentence but, continuing calculation of the LSR with the next sentence. This isolates the influence of each sentence while producing a final information energy sum across the entire text. The ordered information energy is calculated from this with a single sort. If the sorting process requires $n \log(n)$ operations this is clearly slower. With per-sentence, the information energy values are averaged over all the sentences.

For these computations the PTC was broken into 523 files of approximately 5000 words each. Breaks were made on sentence boundaries. Each computation was cross-validated with 1000 Monte Carlo trials, training on 50% of the 523 files and testing on the other 50%. Cross-validation is useful for identifying statistics which work well for evaluation of unknowns rather than ordering of given sets.

**Definition.** Let $V$, for vocabulary, be an ordered collection of words and associated frequency counts. Define the average

$$Z(V) = \tfrac{1}{N} \sum_{r=1}^{N} r \cdot freq(V_r)$$

where $freq(V_r)$ is the frequency count of the rank $r$ word in $V$. The notation $Z$ is used to refer to Zipf's Law [5] which states that rank times frequency is nearly constant in observations from certain "natural" systems. Let $V_{ALL}$ be the vocabulary of all the words in a text. Let $V_{STOP}$ be a collection of syntactic glue words, sometimes called stop words, which have little interest by themselves. Finally, define the statistic

$$Z_{RATIO} = \left[ Z(V_{ALL}) - Z(V_{STOP}) \right] / Z(V_{ALL}). \tag{3}$$

This statistic will be used in the following two computations.

**Computation 2.** Each text in the PTC has an associated date in years which is the earliest of either the publication date or the date of writing when known. The ordered, $IEo$, and unordered, $IEu$, information energy is used to estimate that date.

Date estimation is cast as a least squares problem with the i[th] row of the data matrix constructed as $\langle 1, \quad IEo_i, \quad IEu_i, \quad IEo_i \cdot IEu_i \rangle$. The constant value 1 is used to generate the correct offset for the dates. The estimated dates are compared with the actual dates and the average error is returned.

Figure 3 shows the average date error versus $\tau$ for the six word impulse functions in Table 1 in the per-document case. $I_6(w)$ is the winner with a minimum around $\tau = 110$.

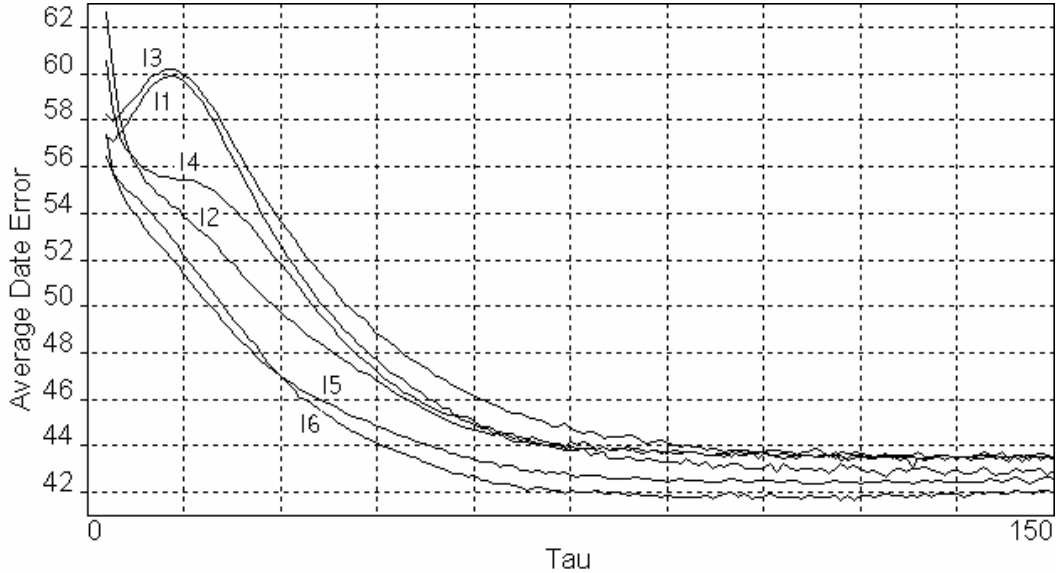

Figure 3. Per-Document Estimation Performance

When the computation was carried out with the test set equal to the training set, $I_1(w)$ was the clear winner at $\tau = 233$ with an error of 37.9 years. This would be the impulse function of choice for ordering a given set.

The per-document case gave better performance than the per-sentence case. In the per-sentence case the error curve could drop to a minimum quickly but would not reach the final performance seen in the per-document case. The best result for the per-sentence case was an average error of 46 years for the $I_3(w)$ function.

When the least squares problem was changed such that the $i^{th}$ row of the data matrix became $\langle 1, \ IEu, \ IEo, \ Z_{RATIO}, \ IEu \cdot IEo, \ IEu \cdot Z_{RATIO}, \ IEo \cdot Z_{RATIO} \rangle$ where $Z_{RATIO}$ is defined in (3), then the average date error was found to be that of Figure 4.

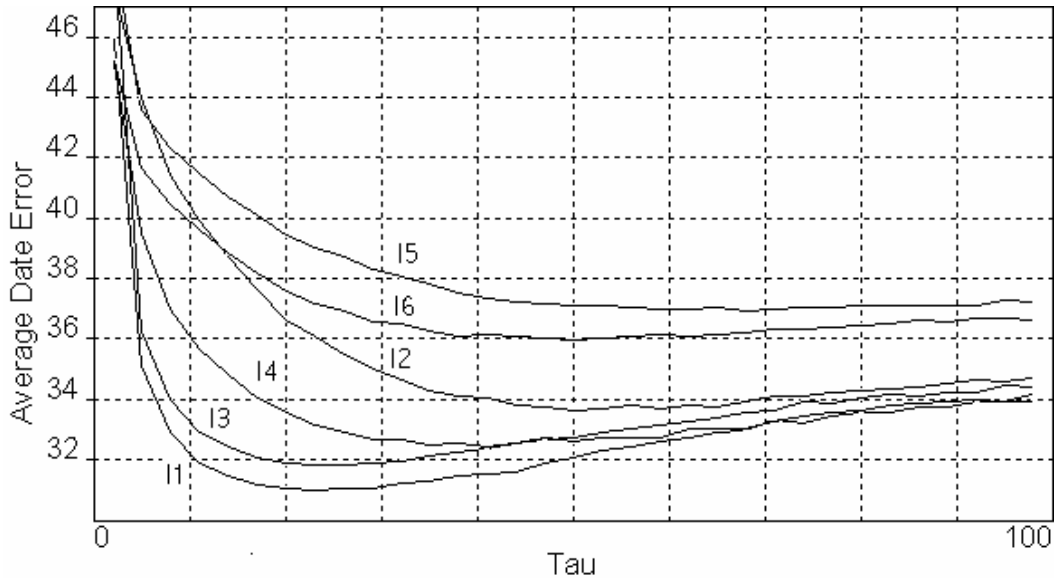
Figure 4.  Per-Document Expanded Estimation Performance

Note that the order of performance is nearly the reverse of Figure 3.  Also note that the location of the minimum for $I_1(w)$ and $I_3(w)$ is for a value of $\tau$ close to the average sentence length.

**Computation 3.**  After breaking the 58 PTC texts into 523 5000 word chunks, the natural question to ask is whether Humpty-Dumpty [6] can be put back together again.  This computation tries to find the best way to do that with information energy.

Here 2-dimensional vectors of the form $vIE = \langle IEo, IEu \rangle$ represent each text and the average vector for each author's training set represents that author.  The minimum distance between a vector from the testing set and any author vector classifies the test text.

Distance is defined in the following way.  For real vectors $X$ and $Y$ and symmetric positive-definite matrix $K$, let $\langle X | Y \rangle_K = X \bullet K \bullet Y$ be an inner product.  Then the distance between two real vectors $U$ and $V$ is $d(U,V) = (\langle U - V | U - V \rangle_K)^{1/2}$.
For this computation K is the inverse of the covariance matrix [7] derived from the set of averaged author training vectors.  Using the inverse covariance matrix has the effect of de-correlating and equalizing the vector coordinates.

Figure 5 shows the performance with respect to $\tau$ of each of the word impulse functions from Table 1 for the per-document case.  $I_5(w)$ and $I_6(w)$ are the winners.  The best performance in the per-sentence case was an error of 0.74.

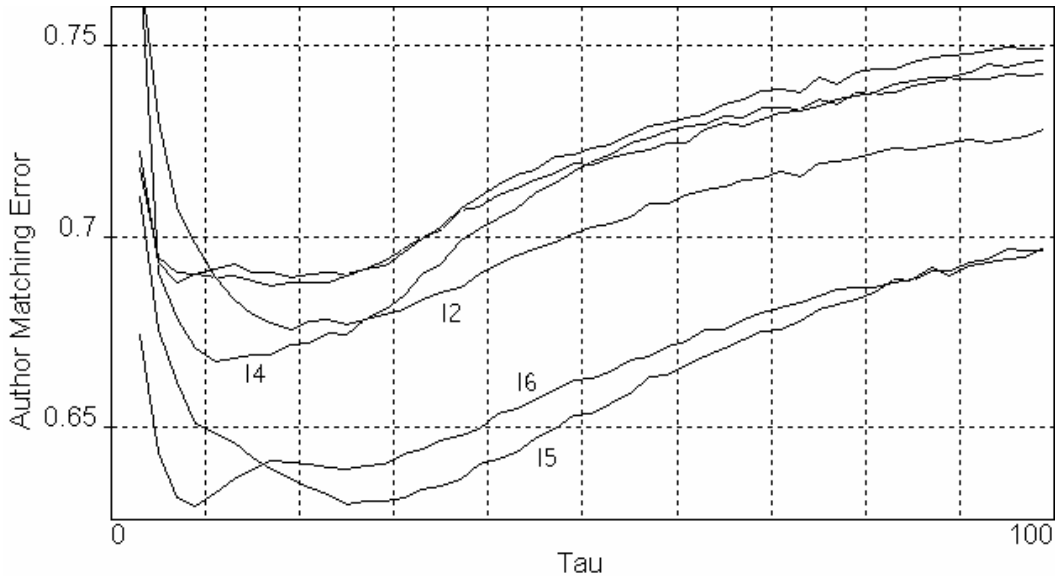
Figure 5.  Per-Document Similarity Performance

When the text representation vector was expanded to $\langle IEo, \quad IEu, \quad Z_{RATIO} \rangle$ the author matching error became that of Figure 6.

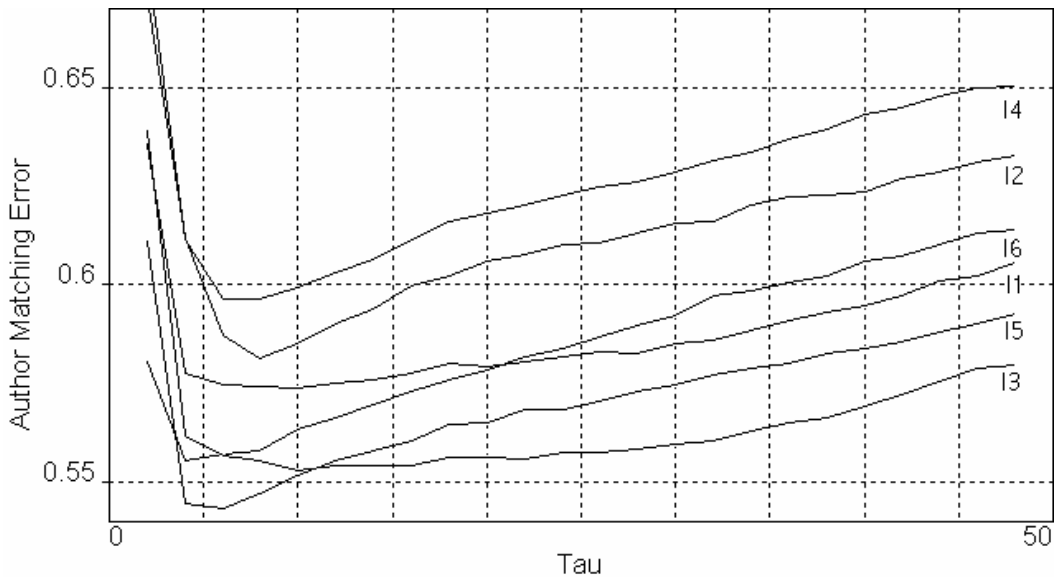
Figure 6.  Per-Document Expanded Similarity Performance

Once again the performance ordering of the word impulse functions has changed with the addition of a new statistic.

**Conclusions.**  The choice of word impulse function and parameter $\tau$ is dependent on circumstance. $I_5(w)$ and $I_6(w)$ work well when information energy is the only statistic.  However, it is expected that it will be used in conjunction with other statistics.  In conjunction with $Z_{RATIO}$, good performance was obtained from the word impulse functions $I_1(w)$ and $I_3(w)$, which have a structure similar or equal to the classic information measure $-\log(\Pr(w))$. It should be noted that the statistic $Z_{RATIO}$ gave the best performance improvement over several standard vocabulary richness statistics tested for this purpose (results not reported here).